\documentclass[journal,twoside,web]{IEEEtran}
\usepackage{generic}
\usepackage{cite}
\usepackage{amsmath,amssymb,amsfonts}
\usepackage{algorithmic}
\usepackage{graphicx}
\usepackage{textcomp}
\usepackage{makecell}

\usepackage[colorlinks=true]{hyperref}
\usepackage{array, booktabs}
\usepackage{balance}
\newcolumntype{P}[1]{>{\centering\arraybackslash}p{#1}}
\newcolumntype{M}[1]{>{\centering\arraybackslash}m{#1}}
\usepackage{adjustbox}
\usepackage{multirow}
\usepackage{float}
\usepackage{threeparttable}
\usepackage{orcidlink}

\usepackage{nomencl}
\makenomenclature

\def\BibTeX{{\rm B\kern-.05em{\sc i\kern-.025em b}\kern-.08em
    T\kern-.1667em\lower.7ex\hbox{E}\kern-.125emX}}
\newcommand{\etal}{\emph{et~al.}}
\DeclareMathOperator{\IN}{IN}
\DeclareMathOperator{\ADAIN}{AdaIN}
\DeclareMathOperator{\SRM}{SRM-FL}
\DeclareMathOperator{\SRMIL}{SRM-IL}

\usepackage{textcomp}
\newcommand\copyrighttext{
  \footnotesize \textcopyright 2024 IEEE. This article has been accepted for publication in IEEE JOURNAL OF BIOMEDICAL AND HEALTH INFORMATICS. See \url{http://www.ieee.org/publications_standards/publications/rights/index.html} for copyright information.}
\newcommand\copyrightnotice{
\begin{tikzpicture}[remember picture,overlay]
\node[anchor=south,yshift=4pt] at (current page.south) {\fbox{\parbox{\dimexpr\textwidth-\fboxsep-\fboxrule\relax}{\copyrighttext}}};
\end{tikzpicture}%
}

\begin{document}

\title{Learning to Generalize towards Unseen Domains via a Content-Aware Style Invariant Model for Disease Detection from Chest X-rays}

\author{Mohammad~Zunaed\textsuperscript{\orcidlink{0000-0001-5987-4800}},~\emph{Student Member,~IEEE}, 
        Md.~Aynal~Haque\textsuperscript{\orcidlink{0009-0003-6750-7623}},~\emph{Senior Member,~IEEE}, and\\ 
        Taufiq~Hasan\textsuperscript{\orcidlink{0000-0002-6142-3344}},~\emph{Senior Member,~IEEE}
\thanks{Manuscript received 22 August 2023; revised 28 December 2023 and 25 February 2024; accepted 29 February 2024. Date of publication 5 March, 2024; date of current version XX XXXXX, 20XX. \textit{(Corresponding author: Taufiq Hasan.)}}
\thanks{Mohammad~Zunaed is with the Department of Electrical and Electronic Engineering (EEE) and also with the mHealth Lab, Department of Biomedical Engineering (BME), Bangladesh University of Engineering and Technology (BUET), Dhaka-1205, Bangladesh. (e-mail: rafizunaed@gmail.com)}
\thanks{Md.~Aynal~Haque is with the Department of EEE, BUET, Dhaka-1205, Bangladesh. (e-mail: aynal@eee.buet.ac.bd)}
\thanks{Taufiq~Hasan is with the mHealth Lab, Department of BME, BUET, Dhaka-1205, Bangladesh. (e-mail: taufiq@bme.buet.ac.bd)}
\thanks{This article has supplementary downloadable material available at \url{https://doi.org/10.1109/JBHI.2024.3372999}, provided by the authors.}
\thanks{Digital Object Identifier \href{https://doi.org/10.1109/JBHI.2024.3372999}{10.1109/JBHI.2024.3372999}}
\\[-0.2ex]
}

\maketitle
\vspace{-10mm}

\begin{abstract}
Performance degradation due to distribution discrepancy is a longstanding challenge in intelligent imaging, particularly for chest X-rays (CXRs). Recent studies have demonstrated that CNNs are biased toward styles (e.g., uninformative textures) rather than content (e.g., shape), in stark contrast to the human vision system. Radiologists tend to learn visual cues from CXRs and thus perform well across multiple domains. Motivated by this, we employ the novel on-the-fly style randomization modules at both image (SRM-IL) and feature (SRM-FL) levels to create rich style perturbed features while keeping the content intact for robust cross-domain performance. Previous methods simulate unseen domains by constructing new styles via interpolation or swapping styles from existing data, limiting them to available source domains during training. However, SRM-IL samples the style statistics from the possible value range of a CXR image instead of the training data to achieve more diversified augmentations. Moreover, we utilize pixel-wise learnable parameters in the SRM-FL compared to pre-defined channel-wise mean and standard deviations as style embeddings for capturing more representative style features. Additionally, we leverage consistency regularizations on global semantic features and predictive distributions from with and without style-perturbed versions of the same CXR to tweak the model's sensitivity toward content markers for accurate predictions. Our proposed method, trained on CheXpert and MIMIC-CXR datasets, achieves 77.32\textpm0.35, 88.38\textpm0.19, 82.63\textpm0.13 AUCs(\%) on the unseen domain test datasets, i.e., BRAX, VinDr-CXR, and NIH chest X-ray14, respectively, compared to 75.56\textpm0.80, 87.57\textpm0.46, 82.07\textpm0.19 from state-of-the-art models on five-fold cross-validation with statistically significant results in thoracic disease classification.
\end{abstract}


\section{Introduction}
\label{Introduction}
\IEEEPARstart{D}{eep} learning algorithms have demonstrated remarkable performance in pathology detection from chest X-ray (CXR) images in recent years \cite{9860074, ZHANG2023102772, JUNG202334, 10106263}. However, despite the promising performance of current deep learning architectures, one major bottleneck is the performance drop on unseen domain data compared to internal data due to distribution discrepancy, also known as domain shift \cite{domain_shift_ref, 10.1109/TPAMI.2022.3195549, KANG2023149, 10.1007/978-3-031-44336-7_17}. This phenomenon hinders the practical applicability of deep learning models for medical image analyses in real-world scenarios. A number of approaches, i.e., domain adaptation (DA) \cite{Luo2020DeepME, ZHANG2022155, 9497656} and domain generalization (DG) \cite{WANG2023104488, 8995481, 9503389, LI2022105144} methods, have been developed over time to address this domain shift.\par
Recent studies have demonstrated that convolutional neural networks (CNNs) exhibit a strong bias toward style rather than content for classification \cite{style_bias_1, geirhos2018}. Geirhos \etal \cite{geirhos2018} showed that CNNs are vulnerable to any changes in image statistics from those on which they are trained. As a result, the CNNs are intrinsically more susceptible to domain shift compared to the human visual system. However, Geirhos \etal \cite{geirhos2018} demonstrated that the heavy texture bias nature of the CNNs can be overcome and tuned toward content bias by utilizing a suitable dataset for training. Jackson \etal \cite{jackson2019style} showed that data augmentation via style transfer with artistic paintings could improve the model's robustness to domain shift for computer vision tasks for natural images. Several DG methods for natural images \cite{zhou2021mixstyle, zhong2022adversarial, 9578071, 9716108} have recently been developed to leverage neural style manipulation for improving cross-domain performances.\par
\begin{figure}[t]
    \centering
	\includegraphics[width=0.95\linewidth]{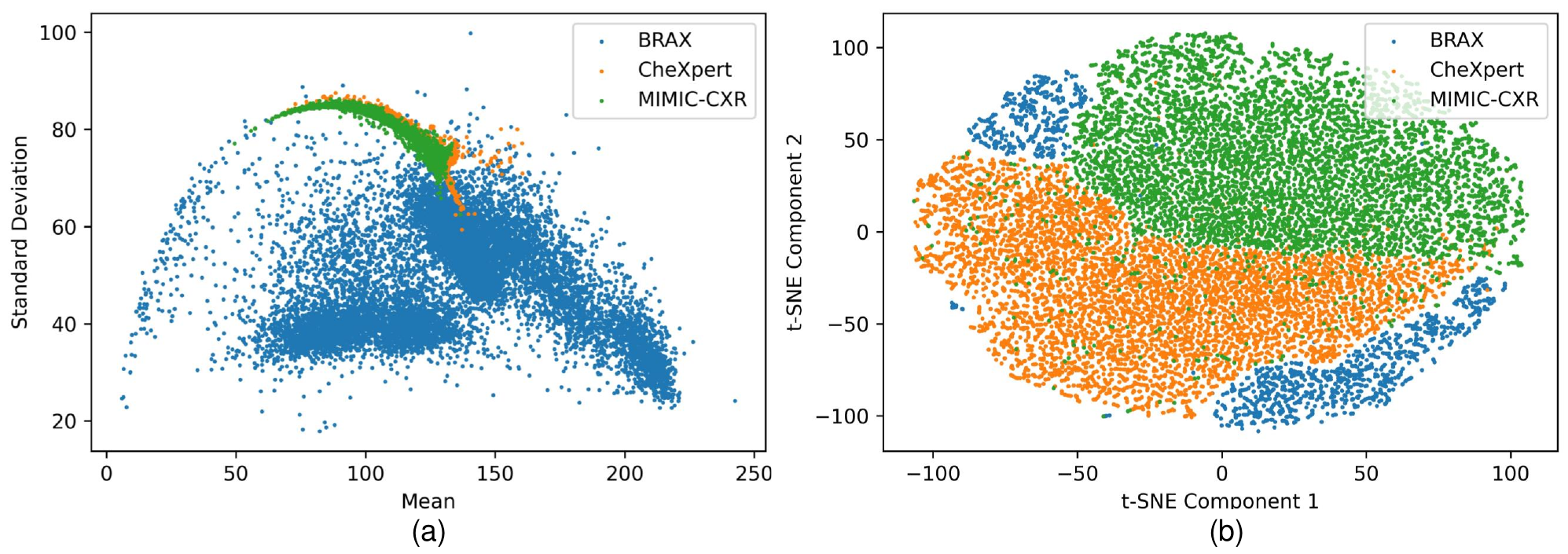}
    \caption{(a) Illustration of image-level distribution gap of three thoracic disease datasets using image-level style embeddings, i.e., mean and standard deviation. (b) 2D t-SNE visualization of style statistics (concatenation of means and standard deviations) computed from the feature maps from the first dense block of the DenseNet-121, trained on the thoracic datasets. We can observe that the feature statistics capture dataset-specific styles reflected by the separable clusters.}
    \label{fig: ds_style_gap_plot}
\end{figure}

\copyrightnotice
Motivated by the impact of neural style randomization on DG for natural images, we experiment with the style statistics of CXR images at both the image and feature levels. First, we plot the mean and standard deviation of CXR images (i.e., image-level style features) from three large-scale thoracic diseases datasets, i.e., CheXpert \cite{chexpert_ds}, MIMIC-CXR \cite{mimic_ds}, and BRAX \cite{brax_ds}. Inspired by the observation from the literature on natural images that feature-level style statistics can often characterize visual domains \cite{zhou2021mixstyle}, we also plot the 2D t-SNE \cite{tsne_ref} visualization of feature-level style statistics (i.e., concatenation of means and standard deviations \cite{adaIN}) computed from the first dense block of DenseNet-121 \cite{8099726} trained on the three thoracic disease datasets. The plots are illustrated in Fig.~\ref{fig: ds_style_gap_plot}. We can observe from Fig.~\ref{fig: ds_style_gap_plot}a that a domain distribution gap based on style statistics exists among these datasets at the image level. The t-SNE plot of the feature statistics in Fig.~\ref{fig: ds_style_gap_plot}b demonstrates that the model captures this uninformative domain information of styles specific to each thoracic disease dataset, which is reflected by the separable clusters. In addition, although CheXpert and MIMIC-CXR are similar at the image level based on simple image statistics, they vary vastly at the feature level because CNNs can extract more high-level and complex features from the images. In real-world scenarios, radiologists tend to learn visual cues from CXRs, rather than uninformative textures and thus perform well across multiple domains. Therefore, in medical imaging for pathology diagnosis from CXR images, robust machine-learning models should not rely on these global image statistics. Instead, they should extract domain-invariant features that are style-invariant and content-biased (pathology traits).\par
However, to the best of our knowledge, no previous study has utilized the application of neural style randomization for domain-agnostic thoracic disease detection from CXR images. DG approaches for natural images based on style randomization \cite{zhou2021mixstyle, nuriel2020padain, 9710533} diversify the training data by synthesizing new styles from the existing sources available during training to simulate unseen domain data. As a result, these methods' augmentation range is sensitive and limited to the style diversity range of the source datasets, which may result in suboptimal performance. In addition, they utilized pre-defined parameters, i.e., mean and standard deviation, as the style embeddings \cite{zhou2021mixstyle, nuriel2020padain, 9716108}. Learnable pixel-wise style embeddings may improve the performance further. Moreover, consistency regularizations between global semantic features and probability distributions of different augmented versions of the same CXR image are not well studied.\par
\copyrightnotice
In this work, we propose a novel domain-agnostic, content-biased, and style-invariant DG framework that can be trained in single or multi-source domain settings without explicit domain labels. We employ style randomization modules (SRMs) at the image and feature levels to hierarchically augment the style while preserving the pathology-specific content traits. Utilizing the prior maximum and minimum pixel value knowledge of a normalized CXR image, we construct a set for sampling mean and standard deviation to use them for switching style characteristics at the image level. For feature-level style augmentation, we deploy a learnable SRM module that generates the pixel-wise style embeddings instead of setting pre-defined parameters, i.e., channel-wise mean and standard deviations as affine parameter values \cite{adaIN}, for style manipulation. In addition, we employ two consistency regularization losses to enforce the consistency of semantic global feature maps and the predicted probability distributions, respectively, between the original CXR and the style-perturbed CXR. The variability of styles between different perturbations of the same CXR with content information intact makes the style attribute unreliable and makes the model focus more on pathology-specific content in making a decision. We also share the source code\footnote{\url{https://github.com/mHealthBuet/DomainAgnosticCXR}} of our method for further dissemination of our approach and experiments. The contributions of this paper are summarized as follows:
\begin{itemize}
    \item We employ a novel SRM at the image level that uses the prior knowledge of possible maximum and minimum pixel values of a CXR to create a set to randomly sample and swap style characteristics.
    \item We propose a novel feature-level SRM module that learns the pixel-level affine parameters instead of utilizing pre-defined style embeddings for creating more rich, diversified style-perturbed features on the fly.
    \item We enforce the consistency regularization losses between the global semantic feature maps and the probability distributions for CXRs with and without style characteristics perturbed to guide the model toward content-related cues.
    \item We perform extensive experiments with five large-scale datasets, i.e., CheXpert \cite{chexpert_ds}, MIMIC-CXR \cite{mimic_ds}, Brax \cite{brax_ds}, NIH chest X-ray14 \cite{nih_ds}, and VinDr-CXR \cite{nguyen2020vindrcxr}. Our proposed framework demonstrates consistent improvements and outperforms competing approaches under different cross-domain settings.
\end{itemize}
\begin{table*}[!t]
\centering
\caption{\textsc{Summary and Comparison of Recent Relevant Methods.}}
\label{table: summary literature review}
\renewcommand{\arraystretch}{1.0}
\begin{adjustbox}{width=\textwidth}
\begin{tabular}{>{\centering}m{0.1\textwidth}>{\centering}m{0.025\textwidth}>{\centering}m{0.5\textwidth}>{\centering\arraybackslash}m{0.3\textwidth}}
\toprule
Articles & Year & Methodology & Limitations\\
\midrule \midrule
FDA \cite{9157228} & 2020 & Utilized the Fourier Transform method to swap low-frequency amplitude components between images, keeping phase components unchanged. & Diversity is limited to the available source domains. Requires access to the target domain.\\
\hline
Luo \etal \cite{Luo2020DeepME} & 2020 & Develop a task-specific adversarial training strategy to reduce the domain discrepancy among CXR datasets.
 & Discriminators require explicit domain labels. \\
\hline
pAdaIN \cite{nuriel2020padain} & 2021 & Swapped style statistics between two instances to create new samples with different style characteristics. & Diversity is limited to the available source domains. Uses pre-defined channel-wise style embeddings. \\
\hline
Tang \etal \cite{9710533} & 2021 & Introduced CrossNorm and SelfNorm modules to swap and recalibrate the feature statistics between two instances to bridge the distribution gap. & Utilizes channel-wise style embeddings. Pixel-wise style embeddings are not considered. \\
\hline
SagNet \cite{9578071} & 2021 & Introduced style and content randomizations along with adversarial style-biased learning to reduce the style bias for making the model more robust under domain shift. & Diversity is limited to the available source domains. Uses pre-defined channel-wise style embeddings. Style-Biased module increases training time.\\
\hline
MixStyle \cite{zhou2021mixstyle} & 2021 & Utilized a parameter-less module to synthesize new style embeddings by probabilistically mixing style statistics from two instances. & Diversity is limited to the available source domains. Uses pre-defined channel-wise style embeddings. \\
\hline
AdvStyle \cite{zhong2022adversarial} & 2022 & Adversarially learned the image-level style features for generating diverse stylized images to reduce overfitting on the source domain. & Perturbation diversity depends on the adversarial gradient. \\
\hline
FSR \cite{9716108} & 2022 & Integrated random noise with the original style parameters to generate new diverse style embeddings. & Utilizes learnable channel-wise style embeddings. Pixel-wise style embeddings are not considered.\\
\hline
Wang \etal \cite{WANG2023104488} & 2023 & Used mixup operation on two different domains to construct a virtual domain and learned multiple domain-specific models to utilize ensemble learning. & Requires at least two available domains to create the virtual domain. \\
\bottomrule
\end{tabular}
\end{adjustbox}
\end{table*}
\section{Related work}
\textbf{Domain Adaptation (DA):} Most past DA approaches have been developed to minimize measures of distances between the source domain and the target domain features \cite{9557808, sym15061163}. Recent methods for DA can be categorized into adversarial training-based domain alignment \cite{Luo2020DeepME, ZHANG2022155}, Fourier transform-based methods \cite{9157228}, generative adversarial networks (GANs) based methods (unpaired image-to-image translation into source/target domain) \cite{9793621, mimic_histogram_gan}, high-level feature map alignment based on similarity measurement \cite{9497656, ZHANG2022155}, gradient reversal based alignment methods \cite{10.5555/2946645.2946704}, and others.\par
Luo \etal \cite{Luo2020DeepME} adopted class-wise adversarial training to alleviate feature differences across common categories of two datasets to mine the knowledge from the external thoracic disease dataset to improve the performance of the classifier on the internal thoracic disease dataset. However, the discriminators require explicit domain labels. Zhang \etal \cite{ZHANG2022155} proposed a domain, instance, and perturbation invariance learning-based framework for utilizing unsupervised domain adaptive thorax disease classification. However, this method requires access to the target domain, which is not often available. Diao \etal \cite{mimic_histogram_gan} proposed a histogram-based GAN methodology for DA that captured the variations in global intensity changes by transforming histograms along with individual pixel intensities. This method is adapted at test time, which is often inconvenient. Yang \etal \cite{9157228} employed the Fourier transform to swap the low-frequency amplitude components between the source and target images while keeping the phase components unaltered to diversify the training dataset. However, this augmentation method is more image-based than goal-oriented.\par
A significant limitation of the DA methods is that they require explicit domain labels, i.e., source and target domains or multiple domains, which is often unavailable in most real-world medical applications due to annotation cost or privacy reasons. In addition, most real-world medical datasets usually consist of data from multiple medical sites using various scanners, i.e., ambiguous domain labels. Moreover, they require re-training during application on new target domains, which is inconvenient and impractical for deploying models in real-world medical image analysis applications.\par
\textbf{Domain Generalization (DG):} DG approaches aim to utilize the source domains only during training and generalize toward any unseen target domain without requiring prior access or re-training \cite{9782500}. DG is considered more challenging than DA approaches, as the target domain samples are not exposed during the training phase. DG methods can be categorized into three broad approaches, i.e., augmentation-based methods \cite{WANG2023104488, 8995481, 9503389}, alignment-based methods \cite{10.1007/978-3-030-58607-2_12, SEGU2023109115}, and meta-learning-based methods \cite{LI2022105144, SANKARANARAYANAN202375}. DG methods can be further divided based on the number of source domains utilized during training: multi-source DG and single-source DG. Multi-source DG assumes multiple relevant domains are available during training. Utilizing multiple source domains allows a deep learning model to discover stable patterns across source domains to better generalize to unseen domains. In contrast, the single-source DG is trained with only one source domain and is expected to generalize toward unseen domains.\par
\copyrightnotice
Wang \etal \cite{WANG2023104488} performed the mixup operation of CXRs from multiple domains to create a virtual domain and utilized ensemble learning to improve performance in unseen domains. However, this approach requires at least two available domains to perform the mixup. Zhang \etal \cite{8995481} exploited a series of extensive deep-stacked augmentations during training to simulate the expected domain shift to address DG on unseen data. However, data augmentation-based methods often require carefully selecting augmentation methods to resemble the unseen domain's appearance.\par
Recent literature studies have shown that CNNs demonstrate a strong bias toward style rather than content and are sensitive to any shift from image statistics that they are trained on, limiting their DG capability \cite{style_bias_1, geirhos2018}. Motivated by the impact of neural style transfer on the domain shift, a number of DG methods have been proposed based on neural style randomization. DG methods based on style randomization can be divided into two categories, i.e., feature-level and image-level randomizations.\par 
The feature-level randomization includes the perturbation of feature statistics from intermediate convolutional layers of the backbone model. Nuriel \etal \cite{nuriel2020padain} swapped the style statistics between activations of permuted and non-permuted samples in a mini-batch to reduce the representation of global statistics in the classifier model. However, they merely swap the style statistics, limiting the sample diversity to the available source domains. Zhou \etal \cite{zhou2021mixstyle} proposed a novel parameter-free module MixStyle that synthesizes new domains by probabilistically mixing feature statistics between instances. Nevertheless, insufficient diversity still exists due to reliance on source domains. Nam \etal \cite{9578071} utilized adaptive instance normalization along with adversarial learning to reduce the intrinsic style bias. However, the style-biased network increases training cost and time. Tang \etal \cite{9710533} proposed two novel modules named CrossNorm and SelfNorm for swapping feature statistics and recalibrating the statistics through attention to both diversify and bridge the gap between source and target domains. Though the SelfNorm is dynamic, the CrossNorm is limited to training data. Wang \etal \cite{9716108} developed an encoder-decoder-like architecture named FSR to integrate random noise with the original style parameters to generate new style embeddings. However, they use the pre-defined parameters, i.e., channel-wise mean and standard deviation, as the style embeddings. Instead of pre-defined parameters, pixel-wise learnable style parameters may yield more diverse and hard style statistics, resulting in a more robust DG model.\par
The image-level style randomization-based methods try to diversify the training dataset by constructing hard stylized samples. Zhong \etal \cite{zhong2022adversarial} proposed a novel augmentation approach, named adversarial style augmentation (AdvStyle), to generate diverse stylized images by adversarially learning the image-level style features to reduce overfitting on the source domain and be robust towards style variations of unknown domains. However, the perturbation diversity depends on the adversarial gradient. Yamashita \etal \cite{9503389} introduced a form of data augmentation named STRAP, based on random style transfer with medically irrelevant style sources for improving DG performance. However, this approach requires training an additional model to perform the stylization.\par
Previous image-level and feature-level style randomization-based DG approaches synthesize style or swap style statistics based on the source domains available during training. Thus, the diversity of the augmented samples is limited and sensitive to the available source domains. Furthermore, the exploitation of semantic content consistency between the original and perturbed activation feature maps will allow for robust feature learning and maximize the effectiveness of style augmentation for the out-of-distribution performance of the model.\par
Overall, to the best of our knowledge, the application of neural style randomization for deriving a DG model for thoracic disease detection from CXRs is not well studied. A systematic exploration of the potential of extracting content-specific (pathology-dependent) but style-invariant domain-agnostic features may further improve DG performance. We have provided a summary of related research in Table \ref{table: summary literature review}.
\copyrightnotice
\section{Methodology}
\subsection{Problem Definition}
Domain generalization focuses on training on single or multiple source domains $\{\mathcal{D}_{S1}, \mathcal{D}_{S2}, \cdots\}$  and is expected to generalize toward any arbitrary unseen target domains $\{\mathcal{D}_{T1}, \mathcal{D}_{T2}, \cdots\}$ with different data distributions. The thoracic disease datasets have significant domain shifts, which may be caused by, but not limited to, various acquisition parameters (exposure shift), manufacturer shift (different hardware, scanning protocol), demographic variations, and so on \cite{WANG2023104488, GARCIASANTACRUZ2021102225}. These variations introduce potential confounders and other sources of bias. Following \cite{WANG2023104488}, we test our proposed method on different datasets compared to the training set for cross-domain performance, referred to as the unseen domain test datasets (out-of-distribution) in this manuscript.
\subsection{Preliminaries}
\subsubsection{Instance Normalization}
Ulyanov \etal \cite{8099920} proposed instance normalization (IN) for better stylization performance. Let the activation feature map, $\mathbf{X_1} \in \mathbb{R}^{D \times W \times H}$, where, $H$ and $W$ indicate spatial dimensions, and $D$ is the number of channels. IN of feature matrix $\mathbf{X_1}$ is formulated as,
\begin{align}
    &\mu(X_1)_{i} = \frac{1}{WH}\sum_{j=1}^{W}\sum_{k=1}^{H}(X_1)_{i,j,k}\\
    &\sigma(X_1)_{i} = \sqrt{\frac{1}{WH}\sum_{j=1}^{W}\sum_{k=1}^{H}\Bigl[(X_1)_{i,j,k}-\mu(X_1)_{i}\Bigl]^{2}+\epsilon}\\
    &\IN(\mathbf{X_1}) =\boldsymbol{\gamma}\left(\frac{\mathbf{X_1}-\mu(\mathbf{X_1})}{\sigma(\mathbf{X_1})}\right)  + \boldsymbol{\beta} \label{eq_in}
\end{align}
Here, $i \in \{1,2, \ldots, D\}$, $\boldsymbol{\gamma}, \boldsymbol{\beta} \in \mathbb{R}^{D}$ are learnable affine vectors, and $\epsilon$ is a very small constant for stability.
\begin{figure*}[!t]
	\centering
	\includegraphics[width=0.9\linewidth]{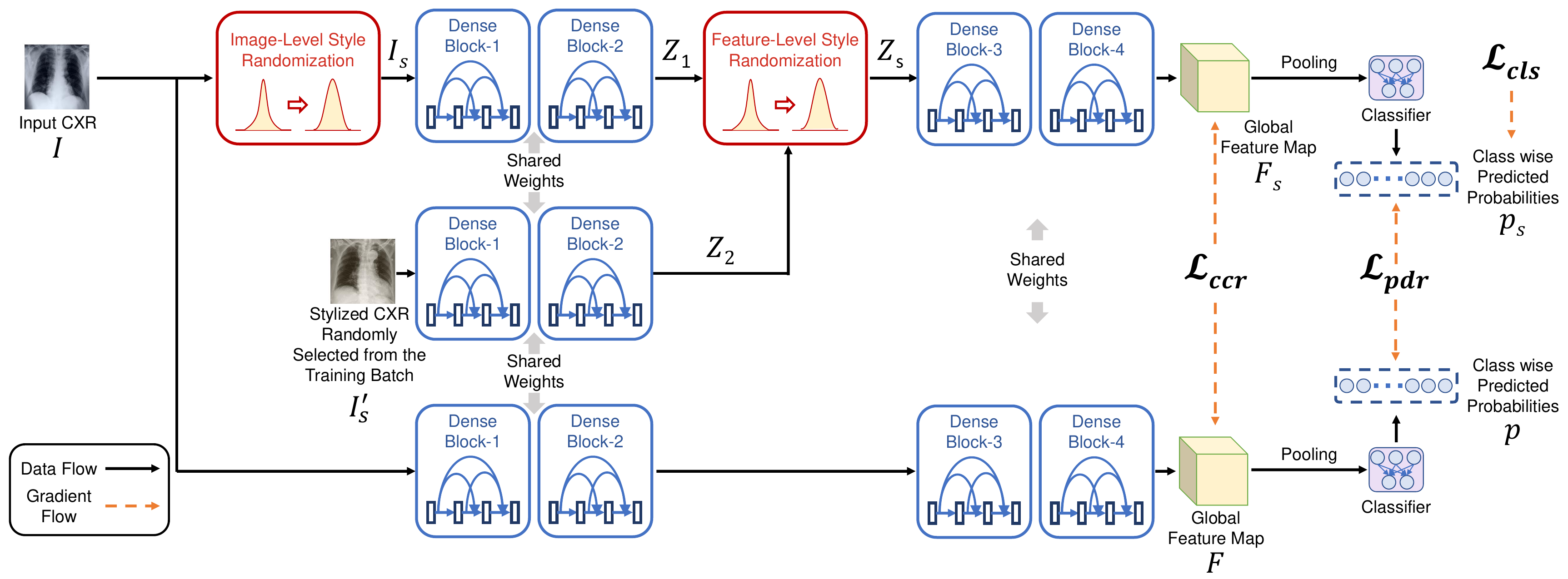}
        \caption{Overview of the proposed framework. The style statistics of the input CXR are randomized with the randomly sampled mean and standard deviation from the set of values constructed from the prior knowledge of possible maximum and minimum values. Both CXRs, with or without style randomized, are passed to the shared feature extractor, DenseIBN-121, to generate two global feature spaces. For the stylized CXR, another round of style randomization is applied to the feature space after dense block-2, with a feature space for a randomly selected stylized CXR from the training mini-batch. The feature-level style randomization block has learnable parameters and is trained alongside the backbone model. While the image-level style randomization block has no learnable parameters, it achieves diversity by sampling different style statistics per CXR image in each iteration for every epoch. The content consistency loss is employed between the two global feature spaces to increase the model's bias toward disease-specific content. In addition, a Kullback-Leibler divergence-based regularization loss is applied between the predicted probability distributions for CXRs with and without style statistics perturbed. The global feature space from the stylized CXR is pooled and passed to the classifier for pathology prediction.}
 	\label{fig: overall_framework}
\end{figure*}
\subsubsection{Adaptive Instance Normalization}
Huang \etal \cite{adaIN} proposed a parameter-less module named adaptive instance normalization (AdaIN) that can change the style of a convolutional feature activation map to any given arbitrary style utilizing the convolutional feature's channel-wise mean and standard deviation as the style statistics. For example, if we want to convert the style of feature map $\mathbf{X_1}$ to that of $\mathbf{X_2} \in \mathbb{R}^{D \times W \times H}$, the channel-wise mean and standard deviation of $\mathbf{X_2}$ will be computed and be used as affine parameters in instance normalization \eqref{eq_in} for stylization.
\begin{align}
    &\ADAIN(\mathbf{X_1}, \mathbf{X_2}) =\sigma(\mathbf{X_2})\left(\frac{\mathbf{X_1}-\mu(\mathbf{X_1})}{\sigma(\mathbf{X_1})}\right) + \mu(\mathbf{X_2}) \label{eq_adain}
\end{align}
In this paper, we will leverage the AdaIN to perturb the style statistics at the image level. For feature-level style perturbation, we will learn the affine parameters for each pixel level for any arbitrary styles instead of using pre-defined channel-wise mean and standard deviation parameters.
\subsubsection{Kullback-Leibler Divergence loss}
The Kullbeck-Leibler divergence ($D_{\text{KL}}(\mathbf{P}||\mathbf{Q})$) loss is applied to minimize the difference between two probability distributions \cite{Joyce2011}, where $\mathbf{P}$ is the target, and $\mathbf{Q}$ is generally the output from a model. In our proposed method, we will derive two predictive distributions, $\mathbf{y_{1}}$, and $\mathbf{y_{2}}$, one for images with style perturbed and the other without style perturbed. As there are no ground truth predictive distributions, to minimize the discrepancy between these predicted predictive distributions, we define the loss as:
\begin{align}
     D_{\text{KL}}^{\prime}(\mathbf{y_{1}}, \mathbf{y_{2}}) &= \frac{D_{\text{KL}}(\mathbf{y_{1}}^*||\mathbf{y_{2}})+D_{\text{KL}}(\mathbf{y_{2}}^*||\mathbf{y_{1}})}{2}
     \label{eq_kld_m}
\end{align}
Note that * indicates that the gradient is not propagated through that parameter, as suggested by Miyato \etal \cite{8417973}, to avoid the model collapse issue.
\subsection{Image-Level Style Perturbation (SRM-IL)}
Let the possible maximum and minimum value of a pixel of a normalized CXR image is $x_{max}$ and $x_{min}$, respectively. Regardless of any domain invariance, the style parameter, such as channel-wise mean and standard deviation, will be limited to this range. For any CXR image $\mathbf{I} \in \mathbb{R}^{w \times h}$, we randomly sample $\mu(\mathcal{S})$ and $\sigma(\mathcal{S})$ from the set of values $\mathcal{S}=[x_{min},x_{max}]$ and use them to transfer the style of the CXR image utilizing AdaIN \cite{adaIN} \eqref{eq_adain}.
\begin{equation}
    \resizebox{.9\hsize}{!}{$\SRMIL(\mathbf{I}, \mu(\mathcal{S}), \sigma(\mathcal{S})) = \mathbf{I_{s}} = \sigma(\mathcal{S})\left(\frac{\mathbf{I}-\mu(\mathbf{I})}{\sigma(\mathbf{I})}\right) + \mu(\mathcal{S})$}
\end{equation}
Here, $h$ and $w$ indicate spatial dimensions. For typical normalized CXR images, the values range from 0 to 255. For other modalities, the maximum and minimum values can be set accordingly. Contrary to the previous methods, we do not utilize existing samples from the source domains for style randomization. As a result, our style diversity is not limited to source domains and can create a more diverse set of global statistics.
\subsection{Feature-Level Style Perturbation (SRM-FL)}
In medical image segmentation, it has been demonstrated that multiple stacked augmentations or strong augmentations can effectively improve a model's robustness to domain shift \cite{8995481}. Inspired by this finding, we apply the style perturbation not only at the image level but also at the feature level. The feature-level style randomization module takes content ($\mathbf{X_{c}} \in \mathbb{R}^{D \times W \times H}$) and reference style ($\mathbf{X_{rs}} \in \mathbb{R}^{D \times W \times H}$) activation maps and transfers the style of the content activation maps to that of the style activation maps. We deploy two style learning nets, i.e., GammaNet ($\phi_{\gamma}$) and BetaNet ($\phi_{\beta}$), to learn the pixel-level affine transformation parameters for transferring style to the content activation maps. The GammaNet and BetaNet shares similar architecture, comprised of four convolutional layers. The initial and the last convolutional layers contain kernels of 1\texttimes1 with the same number of channels as the input feature map. The second and third convolutional layers have filters with 3\texttimes3 dimensions, with channel dimensions following the squeeze and excitation block structure \cite{8578843}, i.e., the second block reduces the channel number, while the third block expands the channel number. The style nets do not contain any non-linear activation functions or normalization layers. The output from these two nets, $\phi_{\gamma}(\mathbf{X_{rs}}) \in \mathbb{R}^{D \times W \times H}$, and $\phi_{\beta}(\mathbf{X_{rs}}) \in \mathbb{R}^{D \times W \times H}$, are set as the affine transformation parameters in the IN function to transfer the style statistics. 
\begin{equation}
    \resizebox{.9\hsize}{!}{$\SRM(\mathbf{X_{c}}, \mathbf{X_{rs}}) = \mathbf{X_{s}} = \phi_{\gamma}(\mathbf{X_{rs}})\left(\frac{\mathbf{X_{c}}-\mu(\mathbf{X_{c}})}{\sigma(\mathbf{X_{c}})}\right) + \phi_{\beta}(\mathbf{X_{rs}})$}
\end{equation}
To train the style nets, we utilize the content and style loss \cite{7780634}. The content and style loss are defined as:
\begin{align}
    \mathcal{L}_{c} &= || \mathbf{X_{c}} - \mathbf{X_{s}} ||_{F}^2 \\
    \mathcal{L}_{s} &= || \mathcal{G}(\mathbf{X_{rs}}) - \mathcal{G}(\mathbf{X_{s}}) ||_{F}^2\\
    \mathcal{L}_{\phi} &= \mathcal{L}_{c} + \eta \mathcal{L}_{s}
\end{align}
Here, $||\cdot||_{F}$ denotes the Frobenius norm, and $\mathcal{G}(\cdot)$ represents the Gram matrix. The $\eta$ is a hyperparameter.
\copyrightnotice
\subsection{Domain Agnostic Network}
The overall proposed framework is illustrated in Fig. \ref{fig: overall_framework}. We adopt DenseIBN-121 \cite{denseIBN} as the backbone of our framework \cite{Luo2020DeepME}. Furthermore, we replace the last fully connected (FC) layer with an FC layer with $N$ output neurons, where $N$ denotes the class number. Please note that our proposed approach is architecture-agnostic and can also be extended to other CNN/Transformer-based backbones. Please refer to subsection A of the supplementary material for experimental results on different backbone architectures.\par 
We take a CXR image $\mathbf{I}$ and apply image-level style perturbation to get perturbed CXR image $\mathbf{I_{s}}$. The pathology traits or content information remain the same in both images. The stylized CXR image $\mathbf{I_{s}}$ is passed through the dense block-1,2 to generate the intermediate feature space $\mathbf{Z_1} \in \mathbb{R}^{D \times W \times H}$. Let $\mathbf{Z_{2}} \in \mathbb{R}^{D \times W \times H}$ be the intermediate feature matrix after the dense block-1,2 for a randomly sampled stylized CXR image $\mathbf{I_{s}}^{\prime}$ from the training mini-batch. We apply the SRM-FL on $\mathbf{Z_1}$ and $\mathbf{Z_2}$ to switch the feature statistics of $\mathbf{Z_1}$ with that of $\mathbf{Z_2}$. 
\begin{equation}
    \mathbf{Z_{s}} = \SRM(\mathbf{Z_1}, \mathbf{Z_2})
\end{equation}
Here, $\SRM(\cdot)$ denotes the feature level style randomization module. Afterward, the feature space $\mathbf{Z_{s}} \in \mathbb{R}^{D \times W \times H}$ is passed to the dense block-3,4 to generate the global feature map $\mathbf{F_{s}} \in \mathbb{R}^{D^{\prime} \times W^{\prime} \times H^{\prime}}$. Here, $D^{\prime}$, $H^{\prime}$, and $W^{\prime}$ denote the feature map depth, height, and width, respectively.\par
\textbf{Classification Loss:} The feature space $\mathbf{F_s}$ is pooled and fed to the classifier network, which consists of one FC layer and one sigmoid layer. Let the predicted probability vector $\mathbf{p^{s}} \in \mathbb{R}^{N}$. $N$ is the total number of pathologies. The pathological ground truths are expressed as a N-dimensional label vector, $\mathbf{L}=[l_1,\ldots,l_i,\ldots,l_N]$ where $l_i \in \{0, 1\},\ N = 14$. Here, $l_i$ represents whether there is any pathology, i.e., 1 for presence and 0 for absence. We utilize the focal loss \cite{8237586} as the classification loss due to class imbalance: 
\begin{align}
    \mathcal{L}_{cls} &= -\alpha^t_i (1-p^t_i)^{\gamma^{\prime}} \log(p^t_i)\\
    p^{t}_i &=
    \begin{cases}
      p^s_i & \text{if $l_i$ = 1}\\
      1-p^s_i & \text{otherwise}\\
    \end{cases}\\
    \alpha^{t}_i &=
    \begin{cases}
      \alpha^t & \text{if $l_i$ = 1}\\
      1-\alpha^t_i & \text{otherwise}
    \end{cases}
\end{align}
Here, $i \in \{1,2,\ldots,N\}$. The $\alpha^t$ and $\gamma^{\prime}$ are the balancing and focusing hyper-parameter from the original paper \cite{8237586}.\par
\textbf{Consistency Regularization Loss:} The CXR without stylization $\mathbf{I}$ is fed through the backbone DenseIBN-121 to generate the global feature map $\mathbf{F} \in \mathbb{R}^{D^{\prime} \times W^{\prime} \times H^{\prime}}$. In neural style transfer, the deep high-level semantic feature maps are utilized for driving the stylization model toward content \cite{7780634, adaIN}. In our method, $\mathbf{F}$ and $\mathbf{F_{s}}$ contain high-level semantic features for input $\mathbf{I}$ and $\mathbf{I_{s}}$, respectively. We hypothesize that if the model is biased toward semantic content, which is domain-irrelevant, then the predicted global feature activation map should be consistent if the model is fed with multiple copies of the same CXRs but with different style attributes. The only difference between $\mathbf{I}$ and $\mathbf{I_{s}}$ is the style perturbation, but the content attributes remain the same in both cases. A model that is biased towards pathological content should be robust to these variations in styles and generate a consistent global feature space. Inspired by the content loss from the literature \cite{7780634, adaIN}, we apply a content consistency regularization loss between these two global feature spaces to tweak the model's sensitivity towards content.
\begin{equation}
    \mathcal{L}_{ccr} = || \mathbf{F_{s}} - \mathbf{F} ||_{F}^2
\end{equation}
where, $||\cdot||_{F}$ denotes the Frobenius norm. In addition, we also utilize the Kullback-Leibler divergence loss for regularizing the predictive distributions from the classifier. Let $\mathbf{p}$ be the predictive distribution vector from the classifier for the CXR image without style perturbation ($\mathbf{I}$). We define the predictive distribution regularization loss, using \eqref{eq_kld_m}, as:
\begin{equation}
    \mathcal{L}_{pdr} = D_{\text{KL}}^{\prime}(\mathbf{p^{s}}, \mathbf{p})
\end{equation}
Thus, the total consistency regularization loss is defined as:
\begin{equation}
    \mathcal{L}_{cons} = \frac{\mathcal{L}_{ccr} + \mathcal{L}_{pdr}}{2}
\end{equation}
The overall learning objective for the backbone model (without style nets) is expressed as:
\begin{equation}
    \mathcal{L}_{total} = \mathcal{L}_{cls} + \mathcal{L}_{cons}
\end{equation}
\begin{table}[!t]
\centering
\caption{\textsc{Training Hyperparameter Settings}}
\label{table: training hyperparameters}
\begin{adjustbox}{width=\linewidth}
\begin{tabular}{ l  c }
\toprule
Training configs & Values \\
\midrule \midrule
Weight initialization& pre-trained on ImageNet\\
Optimizer & Adam \\
Optimizer momentum & $\beta_1$=0.9, $\beta_2$=0.999 \\
Learning rate & 0.0001 \\
Batch size & 50 \\
Gradient accumulation step & 4 \\
Training epochs & 50 \\
Input image dimension & 224\texttimes224 \\
Augmentations & Horizontal flip and Random cropping \cite{9860074}\\
Classification Loss & Focal loss with $\alpha^t$=0.25 and $\gamma^{\prime}$=2 \cite{8237586}\\
Exp. Mov. Avg. (EMA) \cite{ema_cite} & 0.997 \\
\bottomrule
\end{tabular}
\end{adjustbox}
\end{table}

\begin{table*}[!t]
\centering
\caption{\textsc{Performance Comparison with State-of-the-art Systems on the Unseen Domain Test Datasets, i.e., BRAX, NIH Chest X-ray14, and VinDr-CXR, Under the Multi-source Domain Generalization Setting. The Best Result is Shown in \textcolor{red}{Red}.}} 
\label{table: MSDG results}
\begin{adjustbox}{width=\textwidth}
\begin{threeparttable}[b]
\begin{tabular}{*{10}{c}}
\toprule
\multirow{6}{*}{Methods} & \multicolumn{3}{c}{BRAX} & \multicolumn{3}{c}{NIH chest X-ray14} & \multicolumn{3}{c}{VinDr-CXR} \\
\cmidrule(lr){2-4}\cmidrule(lr){5-7}\cmidrule(lr){8-10} & \makecell{BRAX \\ $\rightarrow$ \\ BRAX (ID)} & \multicolumn{2}{c}{\makecell{CheXpert \& MIMIC-CXR \\ $\rightarrow$ \\ BRAX (OOD)}} & \makecell{NIH chest X-ray14 \\ $\rightarrow$ \\ NIH chest X-ray14 (ID)} & \multicolumn{2}{c}{\makecell{CheXpert \& MIMIC-CXR \\ $\rightarrow$ \\ NIH chest X-ray14 (OOD)}} & \makecell{VinDr-CXR \\ $\rightarrow$ \\ VinDr-CXR (ID)} & \multicolumn{2}{c}{\makecell{CheXpert \& MIMIC-CXR \\ $\rightarrow$ \\ VinDr-CXR (OOD)}} \\
\cmidrule(lr){2-2}\cmidrule(lr){3-4}\cmidrule(lr){5-5}\cmidrule(lr){6-7}\cmidrule(lr){8-8}\cmidrule(lr){9-10} & AUC (\%)$^\dagger$ & AUC (\%)$^\dagger$ & p-value$^\ddagger$ & AUC (\%)$^\dagger$ & AUC (\%)$^\dagger$ & p-value$^\ddagger$ & AUC (\%)$^\dagger$ & AUC (\%) & p-value$^\ddagger$ \\
\midrule \midrule
Base & 78.02\textpm0.86 & 73.08\textpm0.75 & 0.0011* & 84.60\textpm0.10 & 81.12\textpm0.13 & $<$0.0001* & 93.09\textpm0.24 & 84.75\textpm0.27 & $<$0.0001* \\
Luo \etal \cite{Luo2020DeepME} & - & 74.63\textpm0.93 & 0.0070* & - & 81.67\textpm0.10 & 0.0004* & - & 86.37\textpm0.47 & 0.0003* \\
Tang \etal \cite{9710533} & 82.79\textpm0.50 & 74.72\textpm0.87 & 0.0042* & 85.01\textpm0.19 & 81.33\textpm0.13 & 0.0002* & 93.90\textpm0.50 & 86.02\textpm0.43 & 0.0007* \\
FDA \cite{9157228} & 83.66\textpm0.41 & 74.93\textpm0.82 & 0.0059* & 85.23\textpm0.10 & 81.47\textpm0.19 & 0.0013* & 94.93\textpm0.26 & 86.75\textpm0.38 & 0.0005* \\
FSR \cite{9716108} & - & 75.25\textpm0.71 & 0.0131* & - & 82.07\textpm0.19 & 0.0031* & - & 87.57\textpm0.46 & 0.0114* \\
Wang \etal \cite{WANG2023104488} & - & 75.33\textpm0.64 & 0.0040* & - & 81.67\textpm0.07 & $<$0.0001* & - & 87.08\textpm0.12 & 0.0003* \\
pAdaIN \cite{nuriel2020padain} & 83.15\textpm0.57 & 75.37\textpm1.01 & 0.0231* & 85.27\textpm0.19 & 81.69\textpm0.04 & 0.0002* & 94.92\textpm0.16 & 86.47\textpm0.43 & 0.0001* \\
MixStyle \cite{zhou2021mixstyle} & 83.31\textpm0.62 & 75.53\textpm0.63 & 0.0069* & 85.34\textpm0.18 & 81.79\textpm0.08 & 0.0007* & 95.01\textpm0.27 & 86.18\textpm0.40 & $<$0.0001* \\
SagNet \cite{9578071} & 83.09\textpm0.69 & 75.48\textpm0.87 & 0.0167* & 85.32\textpm0.26 & 81.76\textpm0.06 & 0.0003* & 95.05\textpm0.16 & 86.56\textpm0.42 & $<$0.0001* \\
AdvStyle \cite{zhong2022adversarial} & 83.23\textpm0.63 & 75.56\textpm0.80 & 0.0139* & 85.19\textpm0.21 & 81.63\textpm0.08 & 0.0002* & 94.89\textpm0.16 & 86.77\textpm0.38 & 0.0002* \\
Ours & 83.31\textpm0.30 & \textcolor{red}{77.32\textpm0.35} & REF & 85.39\textpm0.06 & \textcolor{red}{82.63\textpm0.13} & REF & 95.15\textpm0.21 & \textcolor{red}{88.38\textpm0.19} & REF \\
\bottomrule
\end{tabular}
\begin{tablenotes}
    \item Arrows indicate train data $\rightarrow$ test data, e.g., CheXpert \& MIMIC-CXR $\rightarrow$ BRAX means training on CheXpert \& MIMIC-CXR and
    testing on BRAX.
    \item Abbreviations: ID, in-distribution; OOD, out-of-distribution.
    \item[$\dagger$] Average AUC with the standard deviation of models obtained via cross-validation setting.
    \item[$\ddagger$] A paired t-test on mean AUCs of 5-folds is used for computing statistical significance compared to the reference method (REF).
    \item[*] Indicates a significant difference.
    \end{tablenotes}
    \end{threeparttable}
\end{adjustbox}
\end{table*}

\section{Experiments and Results}
\subsection{Thoracic Disease Datasets}
We utilize the five large-scale thoracic disease datasets, namely, CheXpert \cite{chexpert_ds}, MIMIC-CXR \cite{mimic_ds}, BRAX \cite{brax_ds}, NIH chest X-ray14 \cite{nih_ds}, and VinDr-CXR \cite{nguyen2020vindrcxr} datasets, for our experiments. The CheXpert, MIMIC-CXR, and BRAX datasets have the same label space consisting of 14 pathologies. We follow the U-Ones policy \cite{chexpert_ds} to prepare the ground truths. We use frontal CXR images and select one image per patient/study for our experiments. The VinDr-CXR and NIH chest X-ray14 datasets have 10 and 7 common labels, respectively, with the CheXpert, MIMIC-CXR, and BRAX datasets. We have utilized these common labels for reporting the performance of these two datasets \cite{WANG2023104488}. We use CheXpert and MIMIC-CXR datasets for training and the BRAX, NIH chest X-ray14, and VinDr-CXR datasets for unseen domain inference.
\copyrightnotice
\subsection{Training Scheme}
We resize the CXR images to 256\texttimes256 and normalize them with the mean and standard deviation of the ImageNet training set. We randomly crop 224\texttimes224 patches during training and use a centrally cropped sub-image of 224\texttimes224 dimensions for validation and inference. We provide the summary of training hyperparameter settings in Table \ref{table: training hyperparameters}. We progressively add each proposed component to the base model and utilize the SRM modules only during training time. We employ the percentage area under the receiver operating characteristic curve (AUC) for performance evaluation \cite{Luo2020DeepME, WANG2023104488}. For comparison with other methods, we utilize the multi-label stratified k-fold scheme \cite{iter_strat_cite} to split the training dataset into a 5-fold cross-validation framework to report the mean and standard deviation of the evaluation metrics across the folds. We use the paired t-test between cross-validated mean AUCs for reporting statistical significance.
\subsection{Comparison with the State-of-the-art Methods}
\subsubsection{Multi-source Domain Generalization}
We conduct evaluations and compare with state-of-the-art methods on the BRAX, NIH chest X-ray14, and VinDr-CXR test dataset (unseen domain) and demonstrate the domain generalization capacity of our proposed framework. We compare our proposed method with several existing state-of-the-art models, including the image-level style perturbation-based methods: AdvStyle \cite{zhong2022adversarial}; feature-level style perturbation-based methods: FSR \cite{9716108}, SagNet \cite{9578071}, MixStyle \cite{zhou2021mixstyle}, pAdaIN \cite{nuriel2020padain}, and the method of Tang \etal \cite{9710533}; and Fourier transformation based method: FDA \cite{9157228}. We also compare our method with approaches developed explicitly for thoracic diseases, such as the task-specific adversarial learning approach from Luo \etal \cite{Luo2020DeepME} and the multiple-domain mixup and ensemble learning approach from Wang \etal \cite{WANG2023104488}.\par
We have implemented the methods in the same domain generalization scenario using the same backbone model (DenseIBN-121) and common hyper-parameters, such as learning rate, image dimension, and so on. The hyper-parameters specific to the respective approaches are computed using the 5-fold cross-validation results. The results are summarized in Table~\ref{table: MSDG results}. We have computed both in-distribution (ID) and out-of-distribution (OOD) results and, as expected, found a performance discrepancy across all the models for all three datasets. However, our proposed method achieves the highest performance on the unseen domain test datasets compared to other state-of-the-art methods. As the results show, our proposed method achieves an average AUC of 77.32\textpm0.35, 82.63\textpm0.13, and 88.38\textpm0.19 on the BRAX, NIH chest X-ray14, and VinDr-CXR datasets, respectively, in OOD configuration. Our proposed method achieves a 2.23\%, 0.68\%, and 0.92\% relative average AUC improvement compared to the former best methods on these three datasets, respectively. We have also performed statistical comparisons using the paired t-test on the five-fold cross-validation scores, and the improvements are found to be significant ($p<0.05$). These results suggest that the proposed method has the ability to learn more robust, generalizable, pathology-specific, but domain-agnostic visual features compared to the existing approaches. For ID results, our proposed method achieves competitive results compared to the state-of-the-art approaches. Please note that ID results for the method of FSR \cite{9716108}, Luo \etal \cite{Luo2020DeepME}, and Wang \etal \cite{WANG2023104488} are unavailable as they require multi-domain setup (for discriminators and creation of the virtual domain). Our method achieves 83.31\textpm0.30, 85.38\textpm0.05, and 95.15\textpm0.21 on the BRAX, NIH chest X-ray14, and VinDr-CXR datasets, respectively, in ID configuration. Although FDA \cite{9157228} achieves slightly better results on the BRAX dataset in the ID setting, they have a much larger drop in OOD performance compared to our method, demonstrating the strong domain generalization capability of our method.

\begin{table*}[!t]
\centering
\caption{\textsc{Performance Comparison with State-of-the-art Systems on the Unseen Domain Test Datasets, i.e., BRAX, NIH Chest X-ray14, and VinDr-CXR, Under the Single-source Domain Generalization Setting. The Best Result is Shown in \textcolor{red}{Red}.}}
\label{table: SSDG results}
\begin{adjustbox}{width=\textwidth}
\begin{threeparttable}[b]
\begin{tabular}{*{10}{c}}
\toprule
\multirow{6}{*}{Methods} & \multicolumn{3}{c}{BRAX} & \multicolumn{3}{c}{NIH chest X-ray14} & \multicolumn{3}{c}{VinDr-CXR} \\
\cmidrule(lr){2-4}\cmidrule(lr){5-7}\cmidrule(lr){8-10} & \makecell{BRAX \\ $\rightarrow$ \\ BRAX (ID)} & \multicolumn{2}{c}{\makecell{CheXpert \\ $\rightarrow$ \\ BRAX (OOD)}} & \makecell{NIH chest X-ray14 \\ $\rightarrow$ \\ NIH chest X-ray14 (ID)} & \multicolumn{2}{c}{\makecell{CheXpert \\ $\rightarrow$ \\ NIH chest X-ray14 (OOD)}} & \makecell{VinDr-CXR \\ $\rightarrow$ \\ VinDr-CXR (ID)} & \multicolumn{2}{c}{\makecell{CheXpert \\ $\rightarrow$ \\ VinDr-CXR (OOD)}} \\
\cmidrule(lr){2-2}\cmidrule(lr){3-4}\cmidrule(lr){5-5}\cmidrule(lr){6-7}\cmidrule(lr){8-8}\cmidrule(lr){9-10} & AUC (\%)$^\dagger$ & AUC (\%)$^\dagger$ & p-value$^\ddagger$ & AUC (\%)$^\dagger$ & AUC (\%)$^\dagger$ & p-value$^\ddagger$ & AUC (\%)$^\dagger$ & AUC (\%) & p-value$^\ddagger$ \\
\midrule \midrule
Base & 78.02\textpm0.86 & 71.78\textpm1.46 & 0.0035* & 84.60\textpm0.10 & 79.32\textpm0.51 & 0.0001* & 93.09\textpm0.24 & 80.86\textpm0.81 & 0.0012* \\
Tang \etal \cite{9710533} & 82.79\textpm0.50 & 73.84\textpm0.37 & 0.0005* & 85.01\textpm0.19 & 80.93\textpm0.40 & 0.0004* & 93.90\textpm0.50 & 82.41\textpm0.53 & 0.0004* \\
FDA \cite{9157228} & 83.66\textpm0.41 & 74.25\textpm0.52 & $<$0.0001* & 85.23\textpm0.10 & 81.36\textpm0.27 & 0.0001* & 94.93\textpm0.26 & 82.78\textpm0.52 & 0.0079* \\
pAdaIN \cite{nuriel2020padain} & 83.15\textpm0.57 & 73.98\textpm0.45 & 0.0005* & 85.27\textpm0.19 & 81.48\textpm0.32 & 0.0004* & 94.92\textpm0.16 & 82.63\textpm0.40 & 0.0053* \\
MixStyle \cite{zhou2021mixstyle} & 83.31\textpm0.62 & 74.09\textpm0.16 & 0.0005* & 85.34\textpm0.18 & 81.62\textpm0.26 & 0.0003* & 95.01\textpm0.27 & 82.77\textpm0.57 & 0.0157* \\
SagNet \cite{9578071} & 83.09\textpm0.69 & 74.02\textpm0.31 & 0.0004* & 85.32\textpm0.26 & 81.51\textpm0.31 & 0.0004* & 95.05\textpm0.16 & 82.65\textpm0.47 & 0.0097* \\
AdvStyle \cite{zhong2022adversarial} & 83.23\textpm0.63 & 74.15\textpm0.30 & 0.0002* & 85.19\textpm0.21 & 81.52\textpm0.28 & 0.0003* & 94.89\textpm0.16 & 82.85\textpm0.34 & 0.0051* \\
Ours & 83.31\textpm0.30 & \textcolor{red}{76.27\textpm0.33} & REF & 85.39\textpm0.06 & \textcolor{red}{83.00\textpm0.04} & REF & 95.15\textpm0.21 & \textcolor{red}{84.32\textpm0.36} & REF \\
\midrule \midrule
\multirow{6}{*}{Methods} & \multicolumn{3}{c}{BRAX} & \multicolumn{3}{c}{NIH chest X-ray14} & \multicolumn{3}{c}{VinDr-CXR} \\
\cmidrule(lr){2-4}\cmidrule(lr){5-7}\cmidrule(lr){8-10} & \makecell{BRAX \\ $\rightarrow$ \\ BRAX (ID)} & \multicolumn{2}{c}{\makecell{MIMIC-CXR \\ $\rightarrow$ \\ BRAX (OOD)}} & \makecell{NIH chest X-ray14 \\ $\rightarrow$ \\ NIH chest X-ray14 (ID)} & \multicolumn{2}{c}{\makecell{MIMIC-CXR \\ $\rightarrow$ \\ NIH chest X-ray14 (OOD)}} & \makecell{VinDr-CXR \\ $\rightarrow$ \\ VinDr-CXR (ID)} & \multicolumn{2}{c}{\makecell{MIMIC-CXR \\ $\rightarrow$ \\ VinDr-CXR (OOD)}} \\
\cmidrule(lr){2-2}\cmidrule(lr){3-4}\cmidrule(lr){5-5}\cmidrule(lr){6-7}\cmidrule(lr){8-8}\cmidrule(lr){9-10} & AUC (\%)$^\dagger$ & AUC (\%)$^\dagger$ & p-value$^\ddagger$ & AUC (\%)$^\dagger$ & AUC (\%)$^\dagger$ & p-value$^\ddagger$ & AUC (\%)$^\dagger$ & AUC (\%) & p-value$^\ddagger$ \\
\midrule \midrule
Base & 78.02\textpm0.86 & 69.31\textpm0.34 & $<$0.0001* & 84.60\textpm0.10 & 77.98\textpm0.21 & $<$0.0001* & 93.09\textpm0.24 & 85.73\textpm0.56 & 0.0003* \\
Tang \etal \cite{9710533} & 82.79\textpm0.50 & 72.50\textpm0.62 & 0.0062* & 85.01\textpm0.19 & 79.23\textpm0.10 & 0.0004* & 93.90\textpm0.50 & 86.79\textpm0.49 & $<$0.0001* \\
FDA \cite{9157228} & 83.66\textpm0.41 & 72.96\textpm0.61 & 0.0233* & 85.23\textpm0.10 & 79.65\textpm0.18 & 0.0009* & 94.93\textpm0.26 & 87.17\textpm0.65 & 0.0002* \\
pAdaIN \cite{nuriel2020padain} & 83.15\textpm0.57 & 72.84\textpm0.43 & 0.0048* & 85.27\textpm0.19 & 79.80\textpm0.11 & 0.0075* & 94.92\textpm0.16 & 86.95\textpm0.45 & $<$0.0001* \\
MixStyle \cite{zhou2021mixstyle} & 83.31\textpm0.62 & 73.10\textpm0.48 & 0.0249* & 85.34\textpm0.18 & 79.94\textpm0.06 & 0.0143* & 95.01\textpm0.27 & 86.81\textpm0.42 & $<$0.0001* \\
SagNet \cite{9578071} & 83.09\textpm0.69 & 72.93\textpm0.57 & 0.0166* & 85.32\textpm0.26 & 79.85\textpm0.08 & 0.0101* & 95.05\textpm0.16 & 86.89\textpm0.53 & $<$0.0001* \\
AdvStyle \cite{zhong2022adversarial} & 83.23\textpm0.63 & 73.12\textpm0.54 & 0.0149* & 85.19\textpm0.21 & 79.79\textpm0.09 & 0.0070* & 94.89\textpm0.16 & 87.10\textpm0.47 & $<$0.0001* \\
Ours & 83.31\textpm0.30 & \textcolor{red}{74.23\textpm0.26} & REF & 85.39\textpm0.06 & \textcolor{red}{80.41\textpm0.20} & REF & 95.15\textpm0.21 & \textcolor{red}{89.55\textpm0.34} & REF \\
\bottomrule
\end{tabular}
\begin{tablenotes}
    \item Arrows indicate train data $\rightarrow$ test data, e.g., CheXpert $\rightarrow$ BRAX means training on CheXpert and
    testing on BRAX.
    \item Abbreviations: ID, in-distribution; OOD, out-of-distribution.
    \item[$\dagger$] Average AUC with the standard deviation of models obtained via cross-validation setting.
    \item[$\ddagger$] A paired t-test on mean AUCs of 5-folds is used for computing statistical significance compared to the reference method (REF).
    \item[*] Indicates a significant difference.
    \end{tablenotes}
    \end{threeparttable}
\end{adjustbox}
\end{table*}

\subsubsection{Single-source Domain Generalization}
A more restrictive setting than that of multi-source DG is that of single-source DG, in which only one source domain is available during training. We can seamlessly extend our framework to single-source DG, as our proposed framework does not explicitly require domain labels, unlike some of the other DG methods. To demonstrate the capability of single-source DG, we train our proposed framework on only the CheXpert or MIMIC-CXR dataset, evaluate its performance on the unseen domain test datasets, and compare it with the latest DG methods. The results are reported in Table \ref{table: SSDG results}. From Table \ref{table: SSDG results}, we can observe that our proposed method achieves superior results compared to existing state-of-the-art methods with statistically significant results. 
\copyrightnotice
\subsection{Analysis of Different Components}
We conduct experiments to evaluate the effect of each proposed component on the unseen domain test datasets, including image-level style perturbation (SRM-IL), feature-level style perturbation (SRM-FL), and consistency regularization losses ($\mathcal{L}_{ccr}$ and $\mathcal{L}_{pdr}$). The results are reported in Table \ref{table: ablation study}. Checkmarks in the table indicate if the specific component is incorporated. As shown in Table \ref{table: ablation study}, although all components improve the model performance, SRM-IL and consistency regularization losses contribute more strongly. This stronger contribution is expected since consistency losses reinforce the feature extractor to focus on the content region, and the SRM-IL module increases diversity.\par
We can also observe that the scores achieved by the combination of content consistency regularization ($\mathcal{L}_{ccr}$) and predictive distribution regularization ($\mathcal{L}_{pdr}$) losses surpass the performance level achieved by using either of the regularization loss alone (indices-viii, ix, x). We hypothesize that while predictive distribution regularization loss enforces consistency on two probability distributions from two different style augmented versions of the same CXR image, it does not guarantee that the backbone model will focus on the same spatial region for predicting the consistent probabilities. In order to address this issue, we employ the content consistency loss in addition to the predictive distribution loss. We can also observe that the content consistency loss results in better performance compared to the predictive distribution loss when they are utilized alone.\par
We also conduct an ablation study on the symmetric design of the predictive distribution regularization loss. The results are reported in Table \ref{table: symmetric design of the predictive distribution regularization loss ablation study}. We can observe that utilizing the symmetric design achieves the best scores (indices-iv) while optimizing only one part does not ensure optimal performance (indices-ii and iii).
\copyrightnotice
\begin{figure*}[!t]
    \centering
    \includegraphics[width=0.9\linewidth]{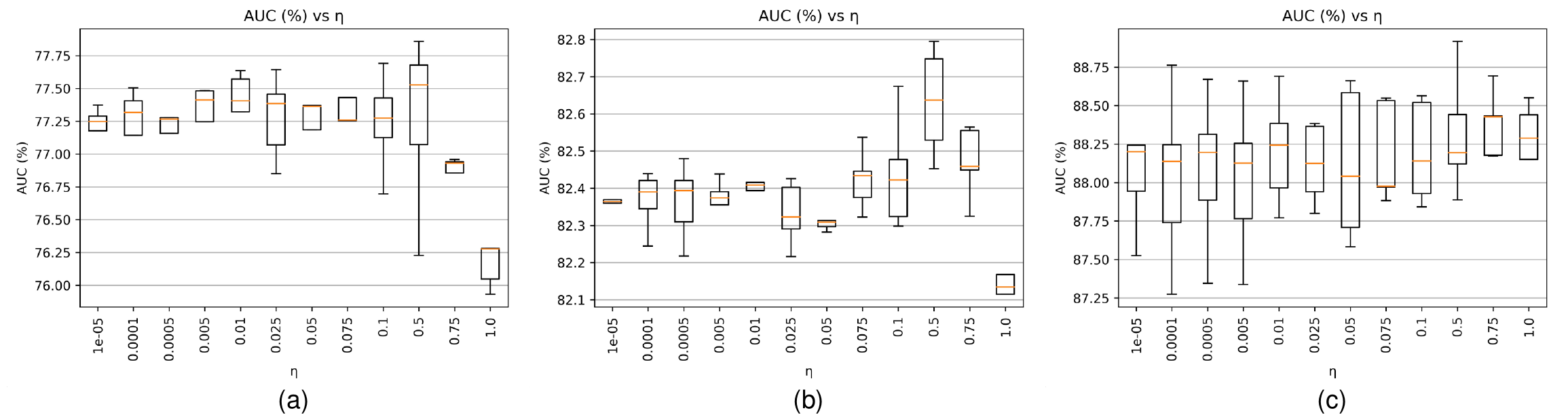}
    \caption{Illustration of the sensitivity of the hyperparameter $\eta$ on (a) BRAX, (b) NIH chest X-ray14, and (c) VinDr-CXR datasets.}
    \label{fig: eta experiment}
\end{figure*}

\subsection{Analysis of SRM-FL} 
\textbf{Effect of randomization stage:} Different layers of the CNN model encode distinct visual attributes, such as lower layers tend to encode low-level texture information while higher levels encode high-level semantic features \cite{7780634, 8977347}. Thus, we must apply the SRM-FL at an adequate stage. This is because employing a SRM-FL to randomize too high-level feature statistics may alter the semantic content necessary for pathology prediction, while randomizing too low-level features may lead to sub-optimal performance. To evaluate the effect of the randomization stage, we apply the SRM-FL after dense block-1,2, and 3, one at a time, and measure the performances. The results are shown in part-1 of Table~\ref{table: feature-level style randomization module ablation}. Our experiment shows that the proposed framework performs best when the SRM-FL is applied after dense block-2.\par
\textbf{Impact of learnable style embeddings:} To evaluate the impact of the learnable pixel-wise affine parameters as the style embeddings compared to the pre-defined style embeddings, we replace the style nets in the SRM-FL with the AdaIN module \cite{adaIN} and compute the percentage AUC score. The results are reported in part-2 of Table~\ref{table: feature-level style randomization module ablation}. We can observe that the learnable pixel-wise affine parameters achieve better results compared to pre-defined style embedding parameters.\par
\textbf{Analysis of the style net architecture:} The style nets $\phi_{\gamma}$ and $\phi_{\beta}$ contain only convolutional (Conv) layers. We conduct experiments adding a rectified linear layer (ReLU) and a batch normalization (BN) layer after the convolutional layers, respectively. The results are reported in part-3 of Table~\ref{table: feature-level style randomization module ablation}. We can observe that the style nets with only convolutional layers achieve the best score. We speculate that the style parameters are set as the affine parameters, and as the affine parameters do not necessarily lie on the positive plane, the utilization of ReLU results in lower performance because it limits the learnable parameter range. In the case of BN, we speculate that the population statistics may interfere with instance statistics, as each single pair (content and style reference sample) has different styles.\par
\copyrightnotice
\begin{table}[!t]
    \centering
    \caption{\textsc{Ablation Study on the Components of Our Proposed Method. The Best Result is Shown in \textcolor{red}{Red}.}}
    \label{table: ablation study}
    \begin{adjustbox}{width=\linewidth}
    \begin{tabular}{ccccccccc}
    \toprule
    \multirow{2}{*}{Indices} & \multirow{2}{*}{Base} & \multirow{2}{*}{SRM-IL} & \multirow{2}{*}{SRM-FL} & \multirow{2}{*}{$\mathcal{L}_{ccr}$} & \multirow{2}{*}{$\mathcal{L}_{pdr}$} & \multirow{2}{*}{BRAX} & \multirow{2}{*}{\makecell{NIH Chest \\ X-Ray14}} & \multirow{2}{*}{VinDr-CXR} \\
    \\
    \midrule
    \midrule
    (i) & \checkmark & & & & & 73.08\textpm0.75	& 81.12\textpm0.13 & 84.75\textpm0.27\\
    (ii) & \checkmark & \checkmark & & & & 76.13\textpm0.62	& 81.91\textpm0.09 & 86.43\textpm0.59\\
    (iii) & \checkmark & \checkmark & \checkmark & & & 76.24\textpm0.51	& 82.03\textpm0.06 & 86.56\textpm0.65\\
    (iv) & \checkmark & & \checkmark & & & 75.19\textpm0.89	& 81.51\textpm0.26 & 86.25\textpm0.40\\
    (v) & \checkmark & \checkmark & & \checkmark & & 77.06\textpm0.29 & 82.46\textpm0.12 & 87.96\textpm0.46\\
    (vi) & \checkmark & \checkmark & & & \checkmark & 76.38\textpm0.40 & 82.30\textpm0.11 & 86.67\textpm0.73\\
    (vii) & \checkmark & \checkmark & & \checkmark & \checkmark & 77.20\textpm0.36 & 82.47\textpm0.06 & 87.68\textpm0.30\\
    (viii) & \checkmark & \checkmark & \checkmark & \checkmark &  & 76.93\textpm0.31 & 82.31\textpm0.04 & 88.37\textpm0.34\\
    (ix) & \checkmark & \checkmark & \checkmark &  & \checkmark & 76.68\textpm0.44 & 82.22\textpm0.08 & 87.43\textpm0.58\\
    (x) & \checkmark & \checkmark & \checkmark & \checkmark & \checkmark & \textcolor{red}{77.32\textpm0.35} & \textcolor{red}{82.63\textpm0.13} & \textcolor{red}{88.38\textpm0.19}\\
    \bottomrule
    \end{tabular}
    \end{adjustbox}
\end{table}
\begin{table}[!t]
\centering
\caption{\textsc{Ablation Study on the Symmetric Design of the $\mathcal{L}_{pdr}$ Loss. The Best Result is Shown in \textcolor{red}{Red}.}}
\label{table: symmetric design of the predictive distribution regularization loss ablation study}
\begin{adjustbox}{width=\linewidth}
\begin{tabular}{ c  c  c  c  c  c }
\toprule
Indices & Optimize y2 & Optimize y1 & BRAX & NIH Chest X-ray14 & VinDr-CXR \\
\midrule \midrule
(i) &   &   & 76.93\textpm0.31 & 82.31\textpm0.04 & 88.37\textpm0.34\\
(ii) &   & \checkmark & 77.00\textpm0.19 & 82.24\textpm0.10 & 87.64\textpm0.44\\
(iii) & \checkmark &   & 76.86\textpm0.24 & 82.32\textpm0.10 & 87.90\textpm0.31\\
(iv) & \checkmark & \checkmark & \textcolor{red}{77.32\textpm0.35} & \textcolor{red}{82.63\textpm0.13} & \textcolor{red}{88.38\textpm0.19}\\
\bottomrule
\end{tabular}
\end{adjustbox}
\end{table}
\begin{table}[!t]
\centering
\caption{\textsc{Analysis of the Feature-Level Style Randomization Module. The Best Result is Shown in \textcolor{red}{Red}.}}
\label{table: feature-level style randomization module ablation}
\begin{adjustbox}{width=\linewidth}
{\renewcommand{\arraystretch}{0.85}
\begin{tabular}{ c  c  c  c }
\toprule
\multicolumn{4}{c}{Part-1: Investigation of the effect of randomization stage} \\
\midrule
Method & BRAX & NIH Chest X-Ray14 & VinDr-CXR \\
\midrule
DB-1 & 77.00\textpm0.39 & 82.34\textpm0.08 & 87.63\textpm0.63 \\
DB-2 & \textcolor{red}{77.32\textpm0.35} & \textcolor{red}{82.63\textpm0.13} & \textcolor{red}{88.38\textpm0.19} \\
DB-3 & 76.75\textpm0.44 & 82.34\textpm0.13 & 88.15\textpm0.35 \\
\midrule
\multicolumn{4}{c}{Part-2: Effect of learnable style embeddings} \\
\midrule
Method & BRAX & NIH Chest X-Ray14 & VinDr-CXR \\
\midrule
Pre-defined & 77.04\textpm0.48 & 82.44\textpm0.13 & 88.09\textpm0.48 \\
Learnable & \textcolor{red}{77.32\textpm0.35} & \textcolor{red}{82.63\textpm0.13} & \textcolor{red}{88.38\textpm0.19} \\
\midrule
\multicolumn{4}{c}{Part-3: Analysis of the style net architecture} \\
\midrule
Method & BRAX & NIH Chest X-Ray14 & VinDr-CXR \\
\midrule
Conv & \textcolor{red}{77.32\textpm0.35} & \textcolor{red}{82.63\textpm0.13} & \textcolor{red}{88.38\textpm0.19} \\
Conv+ReLU & 77.17\textpm0.30 & 82.30\textpm0.10 & 88.23\textpm0.30 \\
Conv+BN & 77.26\textpm0.40 & 82.38\textpm0.14 & 88.19\textpm0.23 \\
\midrule
\multicolumn{4}{c}{Part-4: Effect of style source selection procedure} \\
\midrule
Method & BRAX & NIH Chest X-Ray14 & VinDr-CXR \\
\midrule
Class-balanced Sampling & 77.15\textpm0.29 & 82.39\textpm0.12 & 88.18\textpm0.46\\
Permuted-mini Batch & \textcolor{red}{77.32\textpm0.35} & \textcolor{red}{82.63\textpm0.13} & \textcolor{red}{88.38\textpm0.19}\\
\bottomrule
\end{tabular}}
\end{adjustbox}
\end{table}

\begin{table*}[!t]
    \centering
    \caption{\textsc{Investigation of the Impact of Different Input CXR Dimensions on the BRAX Dataset$^\dagger$. The Best Result is Shown in \textcolor{red}{Red}.}}
    \label{table: spatial dimension eta experiment}
    \begin{adjustbox}{width=\textwidth}
    \begin{threeparttable}[b]
    {\renewcommand{\arraystretch}{0.9}
    \begin{tabular}{cccccccccccccccc} 
    \hline
    \toprule 
    Method & Atel & Card & E.C. & Cons & Edem & Pne1 & Pne2 & P.E. & P.O. & L.L. & L.O. & Frac & S.D. & N.F. & Average \\
    \midrule
    \midrule
    224 \texttimes 224 & 76.78 & \textcolor{red}{86.26} & 73.43 & \textcolor{red}{61.82} & 79.46 & 86.87 & 75.00 & 91.99 & 90.14 & 60.44 & 74.46 & 74.40 & 80.78 & 70.65 & 77.32\textpm0.35\\
    384 \texttimes 384 & \textcolor{red}{78.40} & 86.20 & \textcolor{red}{76.08} & 59.64 & \textcolor{red}{81.57} & \textcolor{red}{88.18} & \textcolor{red}{76.62} & \textcolor{red}{92.55} & \textcolor{red}{90.76} & \textcolor{red}{64.98} & \textcolor{red}{75.32} & \textcolor{red}{76.12} & \textcolor{red}{81.46} & \textcolor{red}{71.62} & \textcolor{red}{78.54\textpm0.29}\\
    \bottomrule 
    \end{tabular}}
    \begin{tablenotes}
    \item[$\dagger$] The 14 findings are Atelectasis (Atel), Cardiomegaly (Card), Enlarged Cardiomediastinum (E.C.), Consolidation (Cons), Edema (Edem), Pneumonia (Pne1), Pneumothorax (Pne2), Pleural Effusion (P. E.), Pleural Other (P.O.), Lung Lesion (L.L.), Lung Opacity (L.O.), Fracture (Frac), Support Devices (S.D.), No Finding (N.F.).
    \end{tablenotes}
    \end{threeparttable}
    \end{adjustbox}
\end{table*}

\textbf{Effect of style source selection procedure:} To evaluate the impact of the style source selection procedure, we construct the style source for each iteration from the training batch by balance sampling of all 14 classes and compare performance with the permuted mini-batch style source selection process. The results are reported in part-4 of Table~\ref{table: feature-level style randomization module ablation}. We can observe that the permuted-mini batch as style source achieves the best score compared to the class-balanced sampling of style source, which can be attributed to the fact that the oversampling of classes with fewer examples results in the model being exposed to fewer style-diverse samples.\par
\begin{figure}[!t]
    \centering
    \includegraphics[width=0.85\linewidth]{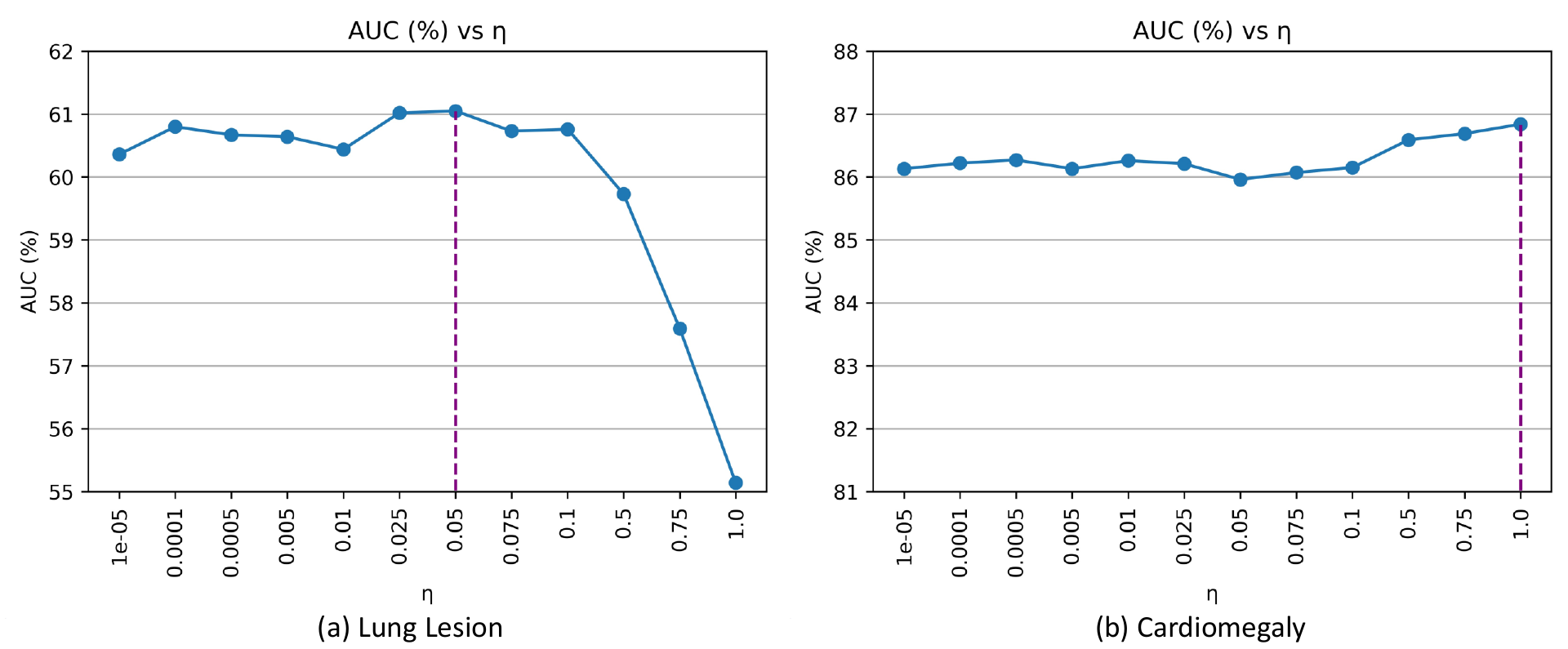}
    \caption{Impact of the hyperparameter $\eta$ on (a) Lung Lesion and (b) Cardiomegaly pathologies (BRAX dataset).}
    \label{fig: disease wise eta experiment (brax)}
\end{figure}

\textbf{Sensitivity of the hyperparameter $\boldsymbol{\eta}$:} We vary the hyper-parameter $\eta$ from 1 to 1e-5 and calculate the percentage AUC performance. The results are demonstrated in Fig.~\ref{fig: eta experiment}. From Fig.~\ref{fig: eta experiment}, we can observe that increasing $\eta$ too much degrades the score as it disrupts the semantic content features. Decreasing the $\eta$ too much reduces the style diversity of the SRM-FL, resulting in performance loss. The hyper-parameter $\eta$ controls the stylization capacity of the style nets by governing the content and style loss. CXRs often have complex backgrounds, and too-hard style perturbations may increase the complexity and potential interference with diagnosing small lesions. To verify this, we plot AUC (\%) scores of two different lesions, i.e., cardiomegaly and lung lesions, compared to the $\eta$ values, on the unseen domain BRAX test dataset. Note that the small lesions, i.e., masses and nodules, are included in the lung lesion category. The plots are given in Fig.~\ref{fig: disease wise eta experiment (brax)}. We can observe that for large lesions such as cardiomegaly, the AUC (\%) does not vary much concerning the $\eta$ value. However, the $\eta$ value significantly impacts the detection of the lung lesion category, i.e., small lesions. Therefore, it is recommended to calculate the optimal $\eta$ value for performing the feature-level style perturbation. Another strategy that can be adopted is to increase the spatial dimension of the input image, which is often utilized in literature for thoracic disease detection to improve the detection of small lesions \cite{9860074}. To demonstrate this, we have trained our model on 384\texttimes384 input images and compared the performance with the 224\texttimes224 input images under $\eta$=0.01, on the unseen domain BRAX test dataset. The results are reported in Table~\ref{table: spatial dimension eta experiment}. We can observe that the AUC (\%) of the lung lesion improves from 60.44 to 65.32.
\copyrightnotice
\section{Discussion \& Conclusion}
In this paper, we propose a DG framework that extracts domain-invariant features by making a deep learning model style invariant and biased toward content. To the best of our knowledge, no previous DG study for thoracic disease detection has exploited neural style transfer to develop a domain-agnostic model. We deploy SRMs both at the image and feature level to augment the style characteristics and employ consistency regularizations to tweak the backbone model's sensitivity towards semantic content rather than uninformative texture, i.e., styles, for pathology prediction. Extensive experiments with five large-scale thoracic disease datasets demonstrate significantly improved performance and generalizability compared to the baseline and state-of-the-art methods on the unseen domain test dataset.\par 
One limitation of our proposed method is that we have to feed both CXR images with or without style attributes perturbed into the backbone to exploit the consistency regularization losses, which increases the training runtime and computing costs. Although our approach has high runtime and computing costs during the offline training stage, it is very efficient for online testing and would be suitable for the routine clinical workflow because the proposed mechanisms are not utilized during the inference stage. Another limitation is that for the SRM-FL, the hyperparameter $\eta$ value needs to be determined through an ablation study, while the SRM-IL and consistency regularization mechanisms are parameterless.\par
Future work directions can involve (1) analyzing how the proposed DG framework can complement the methods designed explicitly for improving thoracic disease detection, e.g., anatomy-aware-based methods, (2) integrating shape-based radiomic features for content regularization, and (3) utilizing local patch statistics instead of global statistics.

\nomenclature[01]{Symbol}{Description}
\nomenclature[02]{$\mu(\cdot)$}{Channel-wise mean}
\nomenclature[03]{$\sigma(\cdot)$}{Channel-wise standard deviation}
\nomenclature[04]{$\epsilon$}{A very small constant for stability}
\nomenclature[05]{$\mathbf{I}$}{Chest X-ray (CXR) image}
\nomenclature[06]{$\mathbf{I_s}$}{Chest X-ray image with style statistics perturbed}
\nomenclature[07]{$\mathbf{X_{1}}, \mathbf{X_{2}}$}{Convolutional feature activation map}
\nomenclature[08]{$\mathbf{X_c}$}{Convolutional feature activation map (content)}
\nomenclature[09]{$\mathbf{X_{rs}}$}{Convolutional feature activation map (style source)}
\nomenclature[10]{$\mathbf{X_{s}}$}{Stylized convolutional activation feature map}
\nomenclature[11]{$\mathbf{Z_1}, \mathbf{Z_2}$}{Intermediate feature activation map for two different image-level stylized CXR images} 
\nomenclature[12]{$\mathbf{Z_{s}}$}{Style perturbed intermediate feature activation map}
\nomenclature[13]{$\mathbf{F}$}{Global feature map}
\nomenclature[14]{$\mathbf{F_s}$}{Global feature map for style perturbed CXR image}
\nomenclature[15]{$\mathcal{S}$}{Style embeddings sampling set for SRM-IL}
\nomenclature[16]{$x_{max}$}{Maximum pixel value of a normalized CXR image}
\nomenclature[17]{$x_{min}$}{Minimum pixel value of a normalized CXR image}
\nomenclature[18]{$\phi_{\gamma}, \phi_{\beta}$}{StyleNets (GammaNet, BetaNet)}
\nomenclature[19]{$\eta$}{Hyperparameter to control the style loss of SRM-FL}
\nomenclature[20]{$\mathbf{P}, \mathbf{Q}$}{Probability Distributions (Target, Predictive)}
\nomenclature[21]{$\mathbf{y_1}, \mathbf{p_s}$}{Predictive distributions for style perturbed CXRs}
\nomenclature[22]{$\mathbf{y_2}, \mathbf{p}$}{Predictive distribution for CXRs without style statistics perturbed}
\nomenclature[24]{$\mathcal{L}_{c}$, $\mathcal{L}_{s}$}{Losses of the StyleNets (Content loss, Style loss)}
\nomenclature[23]{$\mathcal{L}_{cls}$}{Classification loss}
\nomenclature[24]{$\mathcal{L}_{ccr}$}{Content consistency regularization loss}
\nomenclature[25]{$\mathcal{L}_{pdr}$}{Predictive distribution regularization loss}
\nomenclature[26]{$\mathcal{L}_{cons}$}{Total consistency regularization loss}
\printnomenclature[0.43in]

\balance
\bibliographystyle{IEEEtran}
\bibliography{IEEEabrv,main.bib}
\copyrightnotice
\end{document}